\documentclass[runningheads]{eccv_template/llncs}

\usepackage[final,year=2026,ID=3]{eccv_template/eccv}

\usepackage{eccv_template/eccvabbrv}

\usepackage{graphicx}
\usepackage{booktabs}
\usepackage[numbers]{natbib}
\usepackage{listings}
\usepackage{subcaption}
\usepackage{verbatim}
\usepackage{graphicx}
\usepackage{booktabs,longtable}
\usepackage{array}
\usepackage{tabularx}
\usepackage{placeins}
\usepackage{float}
\usepackage{newunicodechar}
\newunicodechar{−}{-}
\usepackage{tabularx}
\usepackage{multirow}
\usepackage{booktabs}
\usepackage{float}
\usepackage{array}
\usepackage{comment}
\usepackage{makecell}
\usepackage{textcomp}
\usepackage{verbatim}
\usepackage{xcolor}
\usepackage{xspace}
\usepackage{pifont}
\usepackage{textcomp}

\usepackage[accsupp]{axessibility}  

\newcommand{\cmark}{\textcolor[HTML]{00B050}{\ding{51}}}
\newcommand{\xmark}{\textcolor[HTML]{FF0000}{\ding{55}}}

\newcommand{\methodname}{\textsc{G3Ego}\xspace}

\usepackage{orcidlink}

\begin{document}


\title{\methodname: Gaze-Guided Graphs for Egocentric Action Understanding}

\titlerunning{Gaze-Guided Graphs for Egocentric Action Understanding}

\author{
Marko Haralovi\'c\inst{1,2}\orcidlink{0009-0004-1178-9964}
\and
Akash Ramakrishnan\inst{2}\orcidlink{0009-0005-1234-2264}\thanks{Corresponding author.}
\and
Estefania Talavera Martinez\inst{2}\orcidlink{0000-0001-5918-8990}
}

\authorrunning{M.~Haralovi\'c et al.}

\institute{
University of Zagreb, Faculty of Electrical Engineering and Computing, Zagreb, Croatia\\
\email{marko.haralovic@fer.hr}
\and
University of Twente, Enschede, The Netherlands\\
\email{a.ramakrishnan@utwente.nl}
}

\maketitle

\begin{abstract}
Egocentric action understanding is often addressed using large video models pretrained on extensive exocentric datasets. However, many first-person actions depend on a small number of hand–object interactions involving only a few relevant entities. 
We propose \methodname, a graph-based framework for egocentric action understanding that uses gaze as a structural cue to identify action-relevant entities in the scene. From sparsely sampled frames, \methodname constructs action scene graphs from vision-language descriptions, grounded objects, and hand cues, and then prunes irrelevant entities using the camera wearer's gaze.
The resulting graph embeddings are temporally aggregated for action recognition and anticipation. Unlike prior work that uses gaze primarily as an auxiliary modality or attention signal, \methodname incorporates gaze directly into graph construction, producing efficient and interpretable representations focused on action-relevant interactions.
Experiments on EGTEA Gaze+ and MECCANO show that \methodname achieves competitive performance compared with video-based approaches and consistently improves Macro-F1 under class-imbalanced evaluation, while avoiding reliance on computationally expensive video pretraining. These results demonstrate the effectiveness of gaze-guided graph representations for egocentric action understanding.
\footnote{\url{https://github.com/MarkoHaralovic/G3Ego}}
\end{abstract}
\section{Introduction}
Egocentric videos captured by wearable devices provide a unique first-person perspective on human behavior and daily activities. They support a wide range of tasks, including egocentric action recognition~\cite{EgoIDT2015CVPR,DeepDescriptors2016CVPR,sudhakaran2018attention,lu2019stam,kapidis2019multitask,mutual_context2020huang,kazakos2021withalittlehelp,wang2021ipl,Li2023Eye,shiota2024egocentric,nasirimajd2025domain}, action anticipation~\cite{furnari2019rulstm,sener2020temp_agg,girdhar2021avt,kini2023meccano,manousaki2023vlmah,mehta2025mmtfru,benavent2026aag,benavent2026aagplus}, gaze anticipation~\cite{zhang2017deepfuturegaze,av_egogaze_anticipation2024,yun2025gazebeyond,ryan2026forecasting}, person re-identification~\cite{basaran2018egoreid}, and human-object interaction~\cite{kapidis2019egocentric,liu2020forecasting,kini2023meccano,thakur2024leveraging,yang2024egochoir,leonardi2026egointeract,materia2026leveraging}.

\begin{figure}[h!]
    \centering
    \includegraphics[width=\columnwidth]{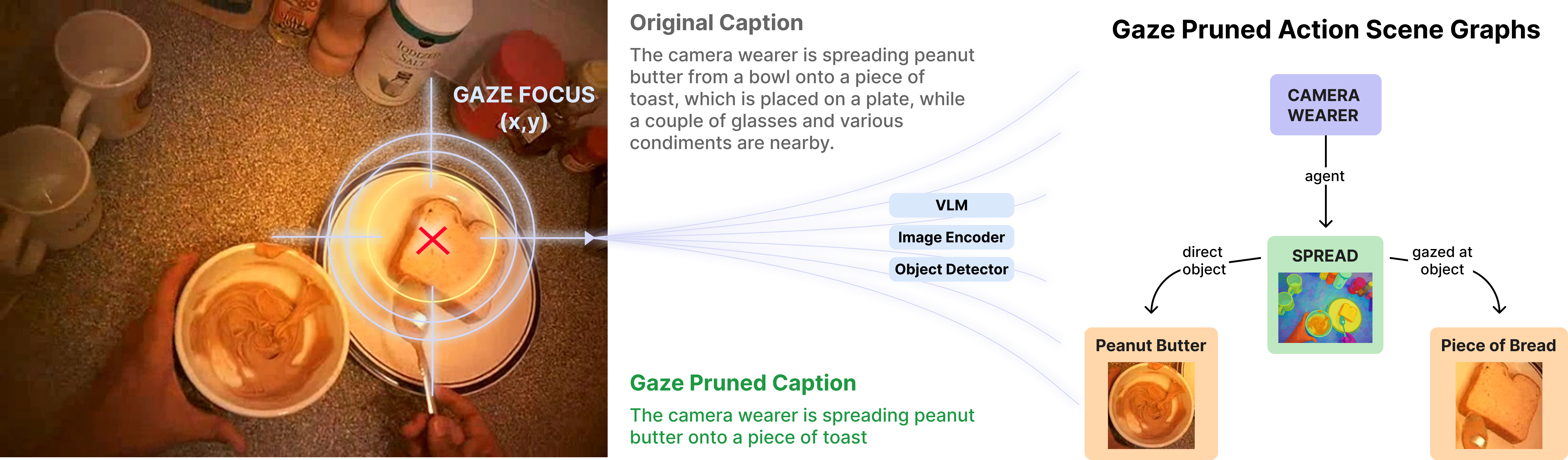}
    \caption{
    \textbf{Motivational overview of gaze-guided graph construction.}
Given an input egocentric frame and the camera wearer's gaze coordinates, a vision-language model generates a caption describing the camera wearer's interactions with the scene. Gaze is used to prune the caption, retaining only action-relevant objects. In this example, the gaze falls on the piece of bread. An action scene graph is then constructed using the global image descriptors from the image encoder as node features for the main verb, and local descriptors from the object detector, including bounding-box features, for the retained objects.}
    \label{fig:motivation_framework}
\end{figure}

Significant progress has been made in exocentric, or third-person, video action recognition~\cite{feichtenhofer2019slowfast,korbar2019scsampler,timesformer2021,smart_frames2021}, and prior work on egocentric action modeling has often relied on video-based models~\cite{EgoIDT2015CVPR,DeepDescriptors2016CVPR,sudhakaran2018attention,sudhakaran2019lsta,lu2019stam,wang2021ipl,furnari2019rulstm,sener2020temp_agg,girdhar2021avt}. While these models are well suited for temporal modeling, they often require large-scale exocentric video pretraining to achieve strong performance~\cite{shiota2024egocentric}, followed by supervised training or fine-tuning on egocentric datasets. Because egocentric videos frequently contain occlusions, partially visible interactions, and abrupt viewpoint changes, visual evidence solely coming from RGB frames may be insufficient for reliable action recognition. Prior work has incorporated complementary input modalities and auxiliary supervision signals. These include multimodal cues~\cite{kazakos2019epicfusion,kazakos2021withalittlehelp,kini2023meccano,hatano2024mmcdfsl,mehta2025mmtfru}, gaze as an auxiliary learning signal~\cite{fathi2012gaze,Li2023Eye,mutual_context2020huang,Min2021GazeAttention,egogazevqa2024,av_egogaze_anticipation2024,gazeaware_detection2024,gazemissteps2025}, and language-based action history or multitask learning, often centered on the camera wearer’s interactions with objects~\cite{kapidis2019multitask,shiota2024egocentric,manousaki2023vlmah,liu2020forecasting,thakur2024leveraging,ni2023hoi_gaze}. Recent progress in high-capacity vision-language models has led to their use in egocentric action modeling and understanding~\cite{lin2022egovlp,pramanick2023egovlpv2,egogazevqa2024,materia2026leveraging,xu2024egoncepp,pei2024egovideo}.

However, most existing methods rely on exocentric video pretraining, egocentric video-domain fine-tuning, and additional supervision, which create computational and annotation requirements for fine-tuning. In this work, we explore graph approaches for egocentric action recognition and anticipation by studying three complementary aspects: (1) leveraging sparsely sampled image-based representations as an alternative to video pretraining, (2) designing graph-based representations for efficient action modeling, and (3) incorporating gaze information as a complementary signal to enhance graph-based modeling.

Inspired by recent graph-based representations, we propose gaze-pruned action scene graphs as an effective structural cue for action recognition and anticipation. Graph-based approaches have been used for action modeling and procedural learning \cite{wang2018spacetime,zhang2019trg,ji2020action,duta2021dynamic,arnab2021unified,li2021videoisgraph,rai2021home,mao2022dynamic,mao2023action,riand2023rethinking,zhao2023constructing,easg2024,benmessabih2025graphbased,taluzzi2026graph}, offering structured inputs and improved interpretability. In our setting, graphs provide a compact alternative to dense visual processing by representing the camera wearer, action predicates, auxiliary verbs, and interacting objects in the scene as nodes, while edges encode semantic relations between objects and actions, as visualized in Figure~\ref{fig:motivation_framework}. We hypothesize that such structured representations capture the essential components of egocentric actions more compactly than frame-based models.

Our automatic pipeline consists of several steps: (a) per-frame scene descriptions, (b) description parsing, (c) frame-based vision encoder inference, (d) object and hand grounding, (e) gaze-guided graph pruning, and (f) graph learning using a temporal aggregation module over graph embeddings. The overall pipeline is shown in Figure~\ref{fig:pipeline}. Since only a small portion of frames contain meaningful action signals, we follow previous work~\cite{easg2024} on sparse clip sampling. Following prior work on gaze as an auxiliary signal~\cite{fathi2012gaze,Li2023Eye,mutual_context2020huang,Min2021GazeAttention,gazemissteps2025}, we introduce a gaze-guided graph pruning strategy that removes non-salient background nodes and retains only objects that the wearer is attending to~\cite{fathi2012gaze}. We refer to these graphs as Gaze-Guided Graphs for egocentric action representation (\methodname).

Prior work has shown that hand- and object-centric cues can be used either as additional model inputs or structured visual cues~\cite{kapidis2019egocentric,kapidis2019multitask,kini2023meccano,shiota2024egocentric,roy2022interaction,pei2025egovideo_hod}, or as auxiliary supervision signals~\cite{liu2020forecasting,thakur2024leveraging,yang2024egochoir,leonardi2026egointeract} for action recognition and human-object modeling. Given their usability, we extract both per-object and per-hand features in a single forward pass and use them as auxiliary visual features encoded in gaze-guided graph object nodes.

Our main contributions are as follows: (a) we introduce \methodname, a gaze-guided graph framework that leverages gaze as a structural cue to prune action scene graphs into compact representations centered on action-relevant entities; and (b) we demonstrate that \methodname achieves competitive performance on egocentric action recognition and anticipation while using substantially more compact representations than dedicated video-based models.
\section{Related Work}
\label{sec:related_work}

\textbf{Action recognition} has been extensively studied using exocentric video benchmarks, with approaches evolving from convolutional architectures~\cite{simonyan2014twostream,carreira2017quo,feichtenhofer2019slowfast} to transformer-based video models~\cite{timesformer2021,liu2022videoswin,yang2023swin3d}. These methods typically rely on large-scale video pretraining datasets, such as Kinetics~\cite{carreira2017quo}, Something-Something-v2~\cite{goyal2017something}, HowTo100M~\cite{miech2019howto100m}, and EPIC-KITCHENS-100~\cite{damen2022epickitchens100}, followed by task-specific fine-tuning.

Egocentric action recognition introduces additional challenges due to the first-person perspective, where actions are often defined by interactions between the camera wearer’s hands and surrounding objects. As a result, existing approaches commonly adapt dense video backbones through egocentric fine-tuning~\cite{ma2016egoconv,EgoIDT2015CVPR,DeepDescriptors2016CVPR,wang2021ipl,pei2024egovideo}, additional supervision signals~\cite{fathi2012gaze,Li2023Eye,Min2021GazeAttention,shiota2024egocentric}, or multimodal inputs~\cite{kazakos2019epicfusion,kazakos2021withalittlehelp,mutual_context2020huang,kapidis2019multitask,manousaki2023vlmah}. More recently, vision-language models have been explored for egocentric representation learning and reasoning~\cite{lin2022egovlp,pramanick2023egovlpv2,pei2024egovideo,pei2025egovideo_hod}. However, these approaches generally continue to rely on large-scale video pretraining and dense temporal representations. In contrast, our work explores whether image-based pretrained representations, combined with structured semantic modeling, can offer an efficient alternative that avoids dedicated video pretraining for egocentric action understanding.

\textbf{Gaze and hand-object cues for egocentric understanding.}
Additional modalities have been widely investigated to address the ambiguity of egocentric video. Among them, gaze provides an important cue about the camera wearer’s attention and has been used to identify task-relevant objects and regions~\cite{fathi2012gaze,EgoIDT2015CVPR}. Subsequent approaches incorporated gaze into attention mechanisms for activity recognition~\cite{Min2021GazeAttention}, jointly modeled gaze and action prediction~\cite{mutual_context2020huang,Li2023Eye,av_egogaze_anticipation2024}, and exploited gaze for human-object interaction understanding~\cite{ni2023hoi_gaze}. Recent works have further explored gaze-guided reasoning for higher-level tasks, including intent understanding with vision-language models~\cite{egogazevqa2024} and interaction anticipation~\cite{materia2026leveraging}.

Hand-object interactions provide another fundamental cue for egocentric action understanding, as many first-person activities are characterized by object manipulation. Previous methods have used hand and object information as explicit inputs~\cite{kapidis2019egocentric,liu2020forecasting,thakur2024leveraging}, auxiliary supervision signals~\cite{shiota2024egocentric,pei2025egovideo_hod}, or cues for reasoning about object states and future interactions~\cite{shiota2024egocentric,ni2023hoi_gaze,materia2026leveraging}. These studies demonstrate that focusing on interaction-relevant entities can reduce the ambiguity caused by irrelevant scene content. Our framework builds upon these observations by using gaze, hands, and objects as structural cues to construct compact action representations.

\textbf{Graph-based representations for action understanding.}
Graphs provide a compact alternative to dense feature representations by explicitly modeling entities and their relationships. Previous works have explored graph-based approaches for video understanding, capturing spatio-temporal relations between regions, frames, or learned visual tokens~\cite{wang2018spacetime,zhang2019trg,duta2021dynamic,arnab2021unified,li2021videoisgraph}. These methods demonstrate the benefits of relational reasoning for action understanding, but typically rely on dense visual backbones and video-level representations.

Scene graphs offer explicit representation of actions by modeling object--entities interactions. Existing approaches have applied scene graphs to compositional action understanding, video question answering, and semantic reasoning~\cite{ji2020action,rai2021home,mao2022dynamic,zhao2023constructing}. However, these methods require dense scene graph annotations, object-centric supervision, or manually designed graph construction procedures, limiting their applicability to unconstrained egocentric scenarios.

More recent work has investigated graph representations for human-centric and egocentric video understanding. Human-centric graphs model relations between people, body cues, and objects~\cite{riand2023rethinking,benmessabih2025graphbased}, while egocentric approaches have explored action scene graphs and graph-based anticipation models~\cite{easg2024,benavent2026aag,benavent2026aagplus}. These methods demonstrate the potential of structured representations for first-person scene understanding, but often depend on additional annotations, explicit temporal alignment, or complex pipelines. In contrast, our method
constructs gaze-pruned action scene graphs directly  from automatically extracted semantic cues and combines them with a lightweight trainable temporal model, without requiring dense video pretraining or manually refined graph supervision.

\section{\methodname: Gaze Guided Graph from Egocentric Videos}
\begin{figure}[h]
    \centering
    \includegraphics[width=.95\linewidth]{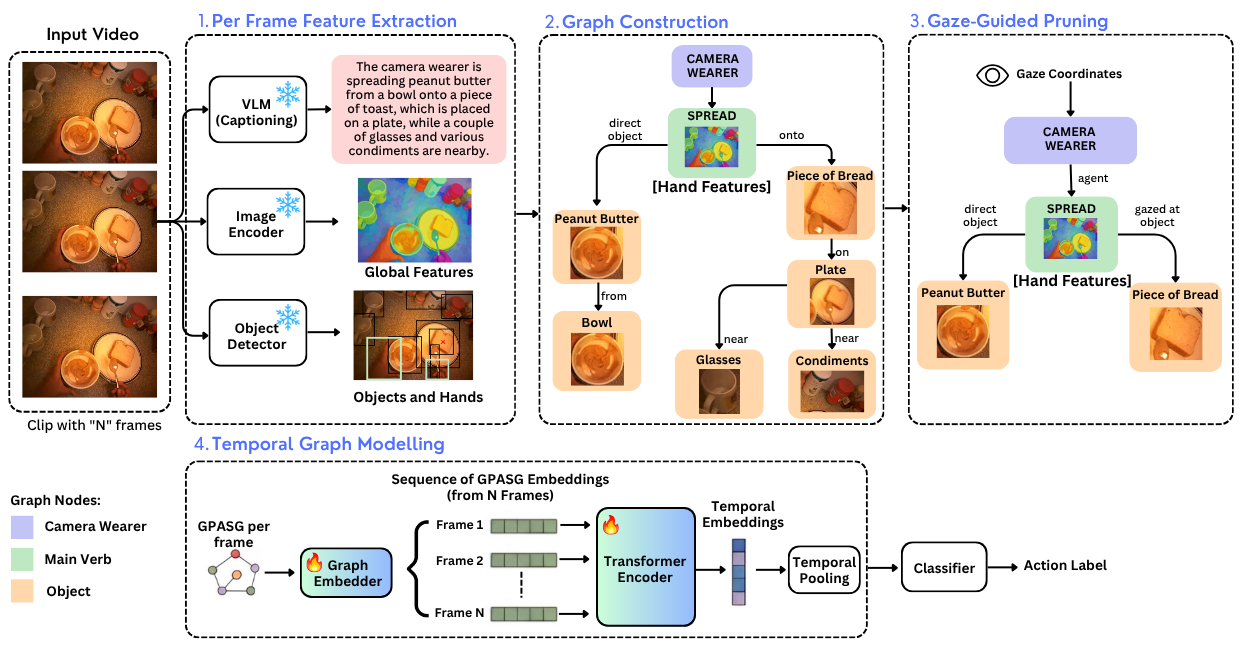}
\caption{\textbf{Overview of the proposed \methodname framework.} Given a clip of $N$ frames, we extract global, object, and hand features, construct a graph of the camera wearer's interactions with the scene, and prune it using gaze. The resulting graph descriptors are temporally aggregated for action recognition.}\label{fig:pipeline}
\end{figure}

We propose \methodname for temporal action understanding in egocentric videos. Given an input video, \methodname constructs a semantic graph representation for each frame by combining global visual descriptors, local object and hand cues, and gaze-guided pruning. The resulting graph sequence is embedded and temporally aggregated for action recognition and anticipation, as illustrated in Fig.~\ref{fig:pipeline}.

\subsection{Frame Representation}
\label{sec:frame_rep}
\paragraph{\textbf{Global Visual and Semantic Descriptors.}}
We extract global frame representations using a frozen vision encoder, producing visual descriptors $\mathbf{x}_t$. Additionally, a Vision Language Model generates a detailed semantic description of each frame, from which verbs, objects, attributes, and relations are extracted using dependency parsing. The exact annotation prompt and decoding configuration are provided in the supplementary material.


\paragraph{\textbf{Local descriptors.}}
To represent local scene elements, we extract objects and their attributes from the parsed annotations. We construct a descriptive list of objects (base object + attributes) and perform visual grounding using an open vocabulary object grounder. This provides both object localization and bounding-box feature vectors, $\mathbf{x}_o$. Each object is assigned a single feature representation. When multiple instances are detected, mean pooling is applied within the frame. To ensure efficiency, we perform a single forward pass per frame and extract feature maps directly, following prior work~\cite{easg2024,air_vqa2025}.

We also extract a 20-dimensional hand feature vector to capture hand location and hand-object interaction cues. For each hand, the representation comprises normalized hand bounding-box coordinates and a side indicator (5D), together with the representation of the interacted object, consisting of its bounding-box coordinates and class index (5D). Features for both hands are concatenated. These features are obtained during the same forward pass used for visual grounding, and a hand is considered to interact with an object when their bounding boxes overlap. If a hand or its interacted object is not detected, the corresponding feature entries are set to zero.

The attributes and relations are encoded as multi-hot matrices of size $|\mathcal{O}|\times|\mathcal{A}|$ and $|\mathcal{O}|\times|\mathcal{R}|$, respectively. During annotation refinement, compound nouns are preserved (e.g., \emph{board game}), while descriptive phrases (e.g., \emph{white doors}) are decomposed into base objects and attribute pairs using dependency parsing. 

\subsection{Graph Construction}
\label{sec:graph_constr}

For each frame $t$, we construct an action scene graph that encodes verbs, objects, their attributes, and semantic relations. Inspired by~\cite{easg2024}, we represent each frame as a graph 
\begin{equation}
    \mathcal{G}_t = (\mathcal{V}_t, \mathcal{E}_t, \mathbf{X}_t),
\end{equation}


where
$
\mathcal{V}_t=\{c,v_t\}\cup\tilde{\mathcal{V}}_t\cup\mathcal{O}_t
$
is the set of graph nodes, consisting of the camera wearer node $c$, the main verb node $v_t$, the set of auxiliary verb nodes $\tilde{\mathcal{V}}_t$, and the set of object nodes $\mathcal{O}_t$. Each node $u\in\mathcal{V}_t$ is associated with a feature vector $\mathbf{x}_u$, collectively denoted by
$
\mathbf{X}_t=\{\mathbf{x}_u \mid u\in\mathcal{V}_t\}.
$
The main verb node is initialized with the global frame descriptor, i.e., $\mathbf{x}_{v_t}=\mathbf{x}_t$, while each object node $o\in\mathcal{O}_t$ is associated with a visual feature vector $\mathbf{x}_o$ and a set of attributes $\mathcal{A}(o)$. The edge set $\mathcal{E}_t$ encodes verb--object interactions (including direct and gazed-at objects), auxiliary--main verb dependencies, and prepositional relations.

The resulting graphs, referred to as \emph{full graphs} (FG), preserve all detected objects, attributes, and semantic relations of the scene. Although comprehensive, they may include entities irrelevant to the performed action, motivating the gaze-guided graph pruning strategy described next.

\subsection{Gaze-Guided Graph Pruning}
\label{sec:gaze_prune}



Unlike previous gaze-guided approaches that primarily use gaze as an auxiliary input or attention signal, our gaze-guided graphs use gaze as a structural cue that determines which entities are preserved in the graph.

\paragraph{\textbf{Gaze-guided graphs.}}

We define gaze-guided pruning as the graph transformation
$
P(\mathcal G_t,\mathbf g_t)=\hat{\mathcal G}_t,
$
where $\mathbf g_t$ denotes the gaze coordinate and
$
\hat{\mathcal G}_t=(\hat{\mathcal V}_t,\hat{\mathcal E}_t,\hat{\mathbf X}_t)
$
is the resulting gaze-guided (pruned) graph.

The gazed object is identified as the object whose bounding-box center is closest to the gaze location,
$
o_t^{g}
=
\arg\min_{o_i\in\mathcal{O}_t}
\left\|
\mathbf{g}_t-\mathbf{c}_i
\right\|_2,
$
where $\mathbf c_i$ denotes the center of the bounding box of object $o_i$. Let
$
\mathcal{O}_t^{v}
=
\{\,o\in\mathcal{O}_t \mid (v_t,o)\in\mathcal{E}_t\,\}
$
denote the set of action-critical objects directly connected to the main verb. The retained node set is then defined as
$
\hat{\mathcal{V}}_t
=
\{c,v_t,o_t^{g}\}
\cup
\mathcal{O}_t^{v}.
$
The corresponding edge set is obtained by restricting the original graph to the retained nodes,
$
\hat{\mathcal{E}}_t
=
\{(u,v)\in\mathcal{E}_t \mid u,v\in\hat{\mathcal{V}}_t\}.
$ The node features $\hat{\mathbf X}_t$ of the pruned graph are inherited from the original graph.

This pruning operation preserves the camera wearer, the main verb, the gazed object, and the action-critical objects while discarding all remaining nodes and edges, producing a compact graph that captures the camera wearer's interactions with the scene.


\subsection{Temporal Graph Modeling}
\label{sec:temporal}

Given a sequence of interaction graphs describing the camera wearer's interactions with the scene, we first encode each graph into a compact descriptor and then model temporal dependencies over the resulting graph representations.

\paragraph{\textbf{Graph Embedder.}}
Each graph is first converted into a tensor representation containing the global visual descriptor, verb and auxiliary verb indices, object representations, object-attribute and object-relation matrices, object features, and a set of (\text{verb}, \text{object}, \text{relation}) triplets.

We then learn a trainable graph embedder that maps these components into a fixed-dimensional representation. Verbs, objects, and relations are embedded using dictionary-based embeddings. Object features and auxiliary verbs are aggregated using Multi-Query Pooling~\cite{lee2019settransformer}, allowing the model to assign different importance to individual elements. The global visual feature is optionally projected into a lower-dimensional space, and the triplets are encoded through an MLP. Finally, we concatenate all embeddings to obtain the graph descriptor.

\paragraph{\textbf{Temporal Aggregation.}}


Given an activity window of $n$ frames, we first construct a sequence of gaze-guided graphs
$
\{\hat{\mathcal G}_1,\hat{\mathcal G}_2,\ldots,\hat{\mathcal G}_n\}.
$
Each graph is mapped by the graph embedder into a $d$-dimensional representation,
$
\mathbf{z}_t=f_{\mathrm{GE}}(\hat{\mathcal G}_t)\in\mathbb{R}^d,
$
where $f_{\mathrm{GE}}(\cdot)$ denotes the graph embedding network.

The resulting sequence is
$
\mathbf Z=[\mathbf z_1,\mathbf z_2,\ldots,\mathbf z_n]\in\mathbb{R}^{n\times d}.
$
To preserve temporal order, a learnable positional embedding
$\mathbf P\in\mathbb{R}^{n\times d}$ is added,
$
\tilde{\mathbf Z}=\mathbf Z+\mathbf P.
$
Sequences shorter than $n$ frames are zero-padded and accompanied by an attention mask, enabling our model to handle variable-length input.

The sequence is then processed by a Transformer encoder
$
\mathbf H=f_{\mathrm{Tr}}(\tilde{\mathbf Z}),
$
where $f_{\mathrm{Tr}}(\cdot)$ denotes a stack of Transformer encoder layers with multi-head self-attention and feed-forward blocks.

The output representations are aggregated using temporal attention pooling,
$
\mathbf h
=
\operatorname{AttnPool}(\mathbf H),
$
producing a fixed-dimensional activity representation. Finally, action probabilities are obtained through a fully connected classifier,
$
\hat{\mathbf y}
=
\operatorname{Softmax}(W\mathbf h+b).
$
\section{Experiments}

\paragraph{\textbf{Datasets.}}
We evaluate our method on two publicly available egocentric datasets on tasks of action recognition and action anticipation, situated in different environments. For both datasets, RGB + gaze is used as input to our method.

MECCANO~\cite{ragusa2021meccano} is an egocentric dataset comprising 20 videos with 8,839 action segments across 61 action types related to the assembly of a toy motorbike. It includes annotations for 20 objects and 12 verbs, contains precomputed object, hands, gaze, and depth features, and provides annotations for both action recognition and action anticipation. We follow the train, validation, and test splits provided by the authors~\cite{ragusa2021meccano}.

EGTEA Gaze+~\cite{Li2023Eye} is an egocentric dataset in a kitchen environment, containing 10,321 segments of 106 actions across three 8:2 train:test splits. It provides hand location annotations and gaze coordinates. Following prior work~\cite{shiota2024egocentric}, we split the training set into train and validation sets so that the final per-split ratio is 7:1:2 for the train, validation, and test sets.
\paragraph{\textbf{Evaluation Details.}} Given an input video sequence of frames $f_1,\ldots,f_n$, our framework operates on sparsely sampled frames and aggregates frame-level predictions to obtain a sequence-level output. We evaluate \methodname on two tasks:

\textit{Action Recognition.}
For action recognition, the goal is to predict the action label associated with the observed frame sequence. We subsample each sequence into $N$ frames, process them independently with our frame-based framework, and temporally aggregate the resulting representations to obtain the final sequence-level prediction.

\textit{Action Anticipation.}
For action anticipation, the goal is to predict the future action label given observations up to timestamp $t$. Specifically, we predict the action occurring at timestamp $t+\delta$, where we set $\delta=1$s in our experiments. We follow the same frame-based processing and temporal aggregation strategy used for action recognition.

\paragraph{\textbf{Implementation Details.}}

We adopt DINOv3~\cite{dinov3_2023} as the frame-based vision encoder, using ViT-L/16 as the backbone. For frame-based VLM captioning, we use Qwen3-VL-32B-Instruct~\cite{bai2025qwen3vl} in mixed precision, with images resized to 448 as input and the maximum number of new tokens set to 160. We rely on spaCy~\cite{honnibal2020spacy} for caption parsing. For object/hand grounding and feature extraction, we use the open vocabulary object grounder GroundingDINO~\cite{groundingdino2023} with a Swin-T backbone, which produces both object features and bounding boxes. Additional details on the pretrained components, feature dimensions, and computational resources are provided in the supplementary material. 

Each graph is constructed from parsed VLM annotations, grounded features obtained with GroundingDINO, hand-object features constructed from groundings, and frame-level semantic features from DINOv3. Each object embedding is a 256-dimensional vector, while DINOv3-L produces a 1024-dimensional vector. Hand features are obtained from the object grounding method, resulting in a 20-dimensional feature vector. 
The resulting graphs are embedded using a lightweight graph encoder into fixed 64-dimensional graph embeddings, which are subsequently processed by the temporal aggregation module adapted from~\cite{sener2020temp_agg}. This module produces a final video representation of 352 dimensions. We compare the proposed model against three baselines: (1) a two-layer bidirectional LSTM with a hidden size of 64, (2) a two-layer MLP with GELU activations and dropout, and (3) a two-layer, four-head Graph Attention Network (GAT) with 128-dimensional node features. While the LSTM and MLP operate on the graph embeddings, the GAT is trained independently as a temporal graph neural network over the sequence of frame graphs. For action recognition, our \methodname model has 105M trainable parameters, while for action anticipation, \methodname has 15M trainable parameters, obtained by reducing the transformer encoder depth and hidden size, as the anticipation involves a shorter effective temporal context.

\paragraph{\textbf{Training Details.}}
Experiments were performed on NVIDIA A40 GPUs with 40GB of VRAM and NVIDIA Quadro RTX 6000 GPUs with 24GB of VRAM, which were used for object grounding and graph training. For VLM inference, we used an NVIDIA RTX PRO 6000 Blackwell GPU with 98GB of VRAM. For training, we use the Adam optimizer with a base learning rate of $3 \times 10^{-4}$ and weight decay of $1 \times 10^{-5}$. We use a linear scheduler with a factor of $0.95$, inverse frequency weighted cross entropy loss, and train for 20 epochs, selecting the best checkpoint based on the macro-F1 score. The batch size is set to 16 and we use 32 frames for graph construction.

\paragraph{\textbf{Offline feature extraction and graph construction.}}
To reduce the computational cost of repeated training and evaluation, we perform the frozen visual feature extraction, both from DINOv3 and GroundingDINO, and graph construction stages once and cache their outputs. The temporal action model is then trained directly on the cached graph representations. Consequently, model training and standard inference over precomputed graphs require only graph embedding and temporal, rather than repeated DINOv3 and GroundingDINO forward passes followed by graph construction, substantially reducing effective computational and memory cost of downstream experimentation.

\paragraph{\textbf{Comparison with State-of-the-Art Methods.}}

On the \textbf{MECCANO action recognition} task, we reproduce the SlowFast baseline~\cite{kini2023meccano} and the recent method of~\cite{benavent2026aag}, since neither report macro-F1 as an evaluation metric. We compare our approach against these reproduced baselines, the ensemble models that achieved the top performance in the MECCANO challenge~\cite{kini2023meccano,tai2022nvidiaunibz}, and the recent method of~\cite{shiota2024egocentric}, which reports F1-score and represents the previous best-performing approach under this evaluation protocol.

For \textbf{action anticipation}, we reproduce existing methods~\cite{sener2020temp_agg,girdhar2021avt,furnari2020rulstm,manousaki2023vlmah,mehta2025mmtfru,benavent2026aag,benavent2026aagplus} and report their top-1 and top-5 accuracy results. Additionally, we include macro-F1 scores for all methods, which we adopt as the primary evaluation metric to account for class imbalance and enable a fair comparison across methods.

For \textbf{EGTEA Gaze+ action recognition}, we compare our method with the methods following two common training protocols: approaches relying exclusively on egocentric fine-tuning~\cite{simonyan2014twostream,EgoIDT2015CVPR,DeepDescriptors2016CVPR,wang2016temporal,sudhakaran2018attention,sudhakaran2019lsta,mutual_context2020huang,wang2021ipl,hao2022group,Li2023Eye}, and approaches leveraging exocentric pretraining followed by egocentric fine-tuning~\cite{carreira2017quo,lu2019stam,kapidis2019multitask,wang2021ipl,shiota2024egocentric,lu2025mixed}. We exclude methods that employ test-time LLM-based filtering~\cite{kazakos2021withalittlehelp,nasirimajd2025domain}, as they introduce additional inference-time information and are therefore not directly comparable under the same evaluation setting.

\paragraph{\textbf{Graph efficiency analysis.}} We analyze the effect of pruning by comparing \methodname with Full Graphs using global efficiency and maximum shortest-path distance. Global efficiency is measured by the following formula 
\begin{equation}
    E_{\mathrm{glob}}(G) = \frac{1}{n(n-1)} \sum_{i \neq j} \frac{1}{d_{ij}}
\end{equation}
where $n$ is the number of nodes and $d_{ij}$ is the shortest path distance between nodes $i$ and $j$. The maximum shortest path from the camera wearer node, denoted  $v_{cw}$, is defined as 
$
    D_{\max} = \max_{v \in \mathcal{V},\, v \neq v_{cw}} d(v_{cw}, v), $
where $d(\cdot,\cdot)$ denotes the shortest path distance. This comparison is done on both datasets.

\section{Results and Discussion}

\begin{table}[t]
\centering
\scriptsize
\setlength{\tabcolsep}{2pt}
\renewcommand{\arraystretch}{1.02}
\caption{Component-wise ablation of \methodname on MECCANO. Global, Object, and Hand denote DINOv3 frame features, Grounding-DINO object-region features, and a 20-dimensional hand/hand--object interaction descriptor, respectively. Gaze is used only for graph pruning, not as a recognition feature. FG denotes the full graph. Macro-F1 is the primary selection metric.}
\label{tab:meccano_design_ablation_study_no_depth}

\begin{tabular*}{\columnwidth}{
    @{\extracolsep{\fill}}
    l c c c c c c c c c @{}
}
\toprule
\makecell{\textbf{Graph}\\\textbf{Type}}
& \makecell{\textbf{Temporal}\\\textbf{Modeling}}
& \makecell{\textbf{N}\\\textbf{Frames}}
& \textbf{Global}
& \textbf{Object}
& \makecell{\textbf{Gaze}\\\textbf{Pruning}}
& \textbf{Hand}
& \textbf{Top-1}
& \textbf{Top-5}
& \textbf{Macro F1} \\
\midrule

\multicolumn{10}{c}{\normalfont\itshape Effect of Object Features} \\[1pt]
FG & LSTM & 10 & \cmark & \xmark & \xmark & \xmark
& 25.58 & 60.75 & 8.29 \\
FG & LSTM & 10 & \cmark & \cmark & \xmark & \xmark
& \underline{37.34} & \underline{70.39} & \underline{8.61} \\

\multicolumn{10}{c}{\normalfont\itshape Effect of Gaze-Guided Graph Pruning} \\[1pt]
FG & LSTM & 10 & \cmark & \cmark & \xmark & \xmark
& 37.34 & 70.39 & 8.61 \\
\methodname & LSTM & 10 & \cmark & \cmark & \cmark & \xmark
& \underline{37.90} & \underline{70.88} & \underline{10.63} \\

\midrule
\multicolumn{10}{c}{\normalfont\itshape Effect of Sampling Rate} \\[1pt]
\methodname & LSTM & 1  & \cmark & \cmark & \cmark & \xmark
& 31.85 & 64.86 & 4.72 \\
\methodname & LSTM & 10 & \cmark & \cmark & \cmark & \xmark
& 37.90 & 70.88 & 10.63 \\
\methodname & LSTM & 32 & \cmark & \cmark & \cmark & \xmark
& \underline{39.50} & \underline{72.33} & \underline{12.34} \\

\midrule
\multicolumn{10}{c}{\normalfont\itshape Temporal Modeling Architecture} \\[1pt]
\methodname & MLP        & 32 & \cmark & \cmark & \cmark & \xmark
& 35.28 & 69.61 & 7.71 \\
\methodname & GNN        & 32 & \cmark & \cmark & \cmark & \xmark
& 33.79 & 67.62 & 8.70 \\
\methodname & LSTM       & 32 & \cmark & \cmark & \cmark & \xmark
& 39.50 & 72.33 & 12.34 \\
\methodname & Temp.\ Agg. & 32 & \cmark & \cmark & \cmark & \xmark
& \underline{41.91} & \underline{79.63} & \underline{15.87} \\

\midrule
\multicolumn{10}{c}{\normalfont\itshape Input Feature Ablation} \\[1pt]
\methodname & Temp.\ Agg. & 32 & \xmark & \cmark & \cmark & \cmark
& 26.99 & 60.64 & 3.83 \\
\methodname & Temp.\ Agg. & 32 & \cmark & \cmark & \cmark & \xmark
& 41.91 & 79.63 & 15.87 \\
\methodname & Temp.\ Agg. & 32 & \cmark & \cmark & \cmark & \cmark
& \textbf{46.48} & \textbf{82.04} & \textbf{21.34} \\

\bottomrule
\end{tabular*}
\end{table}

\paragraph{\textbf{Framework components ablation.}}
We first conduct an ablation study on the MECCANO action recognition dataset \cite{ragusa2021meccano} to compare and validate the contribution of the individual components of our framework. The ablation results in Table \ref{tab:meccano_design_ablation_study_no_depth} show that gaze pruning provides an improvement over the full action scene graph (FG). When using 10 frames with an LSTM backbone for aggregation, \methodname improves Top-1 accuracy from 37.34 to 37.90, Top-5 from 70.39 to 70.88, and Macro-F1 from 8.61 to 10.63 over FG. This indicates that removing irrelevant context helps the model focus on the most action-relevant objects.   Additional experiments examining the VLM semantic prior, random pruning, and explicit gaze input are reported in the supplementary material.

We also observe that temporal context plays an important role. Increasing the number of frames sampled within the same temporal window from 1 to 10, and then to 32, steadily improves the performance, with Top-1 accuracy increasing from 31.85 to 39.50, and Macro-F1 increasing from 4.72 to 12.34. This suggests that performance improves when the model observes a denser sampling of frames within the same temporal window.

We compared different models to aggregate the graph embeddings across all the frames. On \methodname with 32 frames, the MLP baseline reaches a Top-1 accuracy of 35.28 and Macro-F1 of 7.71, GNN obtained a Top-1 of 33.79 and Macro-F1 of 8.70, LSTM achieved a Top-1 of 39.50 and 12.34, and the proposed temporal aggregation model achieved the best performance with a Top-1 accuracy of 41.91 and Macro-F1 of 15.87.

Finally, we experimented with the removal of RGB features, which caused the performance to drop sharply to 26.99 Top-1 and 3.83 Macro-F1, and inclusion of hand features on \methodname with temporal aggregation, which reaches the best performance of 46.48 Top-1, 82.04 Top-5, and 21.34 Macro-F1.

\begin{table}[t]
\centering
\small
\setlength{\tabcolsep}{4pt}
\renewcommand{\arraystretch}{1.05}
\caption{Performance comparison with SOTA on MECCANO action recognition task. AH: action history; HOCL: hand-object contact labels; OSL: object state labels.}
\label{tab:meccano_main_results}
\resizebox{\columnwidth}{!}{%
\begin{tabular}{l l c c c c c}
\toprule
\textbf{Method} & \textbf{Venue} & \textbf{Modality} & \textbf{Top-1} & \textbf{Top-5} & \textbf{Macro-F1} & \textbf{Train.\ Params} \\
\midrule
Majority class baseline & -- & -- & 27.31 & 51.22 & 0.72 & -- \\
\midrule
SlowFast~\cite{kini2023meccano} & CVPRW'23 & RGB & 45.16 & 73.75 & 18.91 & 68M \\
SlowFast~\cite{kini2023meccano} & CVPRW'23 & RGB+Depth+Gaze & 49.66 & 77.82 & -- & 68M \\
LUBECK UniFormer ens.~\cite{kini2023meccano} & CVPRW'23 & RGB+Depth & 51.82 & 83.35 & -- & 540M \\
UCF Swin3D-B~\cite{kini2023meccano} & CVPRW'23 & RGB+Depth & \textbf{52.82} & \textbf{83.85} & -- & 176M \\
UNIBZ 8-model ens.~\cite{bianchi2024egocentric} & ISIEA'24 & RGB+Depth & 52.57 & 81.53 & -- & 194M \\
Swin-B + HOCL+OSL~\cite{shiota2024egocentric} & WACV'24 & RGB+HOCL+OSL & 44.81 & 77.01 & 16.70 & 176M \\
AAG~\cite{benavent2026aag}  & WACV'26 & RGB+Depth+AH & 33.48 & 69.32 & 9.52 & 24M \\
\midrule
\methodname & -- & Gaze(RGB + Hands) & 46.48 & 82.04 & \textbf{21.34} & 105M \\
\bottomrule
\end{tabular}%
}
\end{table}

\paragraph{\textbf{Action recognition performance on MECCANO.}}
Table \ref{tab:meccano_main_results} compares our method with reported state-of-the-art approaches on MECCANO action recognition. Our model achieves 46.48 Top-1, 82.04 Top-5, and 21.34 Macro-F1 using \methodname with RGB and hands, which places it competitively among prior methods. While the best Top-1 accuracy is obtained by UCF Swin3D-B~\cite{kini2023meccano} with 52.82, our method outperforms several other strong baselines, including SlowFast with RGB \cite{kini2023meccano}, Swin-B+HOCL+OCL \cite{shiota2024egocentric}, and AAG \cite{benavent2026aag}. Notably, \methodname achieves the highest reported Macro-F1 score (21.34), indicating improved robustness across action categories.

\begin{table}[t]
\centering
\scriptsize
\setlength{\tabcolsep}{2.4pt}
\renewcommand{\arraystretch}{0.92}
\caption{EGTEA Gaze+ action-recognition results on the official three splits. S1--S3 report mAcc when available; otherwise they report per-split Top-1/overall accuracy. MTL: multi-task learning; IDT:improved dense trajectories.}
\label{tab:egtea_results}
\resizebox{\columnwidth}{!}{%
\begin{tabular}{@{}llp{3.1cm}ccccc@{}}
\toprule
\textbf{Method} & \textbf{Venue} & \textbf{Modality} &
\textbf{S1} & \textbf{S2} & \textbf{S3} & \textbf{Avg mAcc} & \textbf{Avg Top-1} \\
\midrule

Two Stream~\cite{simonyan2014twostream} & NeurIPS'14 & RGB + Optical Flow & 43.78 & 41.47 & 40.28 & 41.84 & -- \\
EgoIDT+Gaze~\cite{EgoIDT2015CVPR} & CVPR'15 & RGB + IDT + Gaze & 42.55 & 37.30 & 37.60 & 39.13 & -- \\
I3D+EgoConv~\cite{DeepDescriptors2016CVPR} & CVPR'16 & RGB + Hands + Head + Gaze & 54.19 & 51.45 & 49.41 & 51.68 & -- \\
TSN~\cite{wang2016temporal} & ECCV'16 & RGB + Optical Flow & \underline{58.01} & \underline{55.01} & \textbf{54.78} & 55.93 & -- \\

Sudhakaran et al.~\cite{sudhakaran2018attention} & ECCVW'18 & RGB & 52.40 & 50.09 & 49.11 & 50.53 & -- \\
LSTA~\cite{sudhakaran2019lsta} & CVPR'19 & RGB + Optical Flow & 53.00 & -- & -- & -- & -- \\
Mutual Context~\cite{mutual_context2020huang} & TIP'20 & RGB + MTL & 55.70 & -- & -- & -- & 62.60\\
I3D~\cite{wang2021ipl} & ICCV'21 & RGB & 56.78 & 54.92 & 53.94 & 55.21 & -- \\
GC-TSM~\cite{hao2022group} & CVPR'22 & RGB & -- & -- & -- & -- & \textbf{65.10} \\
I3D+Gaze~\cite{Li2023Eye} & TPAMI'23 & RGB + Gaze & 53.74 & 50.30 & 49.63 & 51.22 & -- \\
Prob-ATT w/ gaze~\cite{Li2023Eye} & TPAMI'23 & RGB + Gaze Attention & 57.20 & 53.75 & \underline{54.13} & 55.03 & -- \\
\midrule
\methodname & -- & Gaze(RGB + Hands) & \textbf{61.68} & \textbf{56.34} & 51.46 & \textbf{56.49} & \underline{64.25} \\
\bottomrule
\end{tabular}%
}
\end{table}

\begin{table}[t]
\centering
\small
\setlength{\tabcolsep}{4pt}
\renewcommand{\arraystretch}{1.05}
\caption{MECCANO next-action prediction results for reported methods and our graph-based models. Results are reported for action anticipation at $t_a=1$s before the next action. \textbf{Bold} values indicate the best result across the full table.}
\label{tab:meccano_reported_graph_results}
\resizebox{\columnwidth}{!}{%
\begin{tabular}{l l c c c c c}
\toprule
\textbf{Method} & \textbf{Venue} & \textbf{Modality} & \textbf{Top-1} & \textbf{Top-5} & \textbf{Macro-F1} & \textbf{Train.\ Params} \\
\midrule
Majority class baseline & -- & -- & 27.21 & 51.15 & 0.71 & -- \\
\midrule
TempAgg \cite{sener2020temp_agg} & ECCV'20 & OHG & 19.69 & 25.37 & 3.11 & 123M \\
AVT \cite{girdhar2021avt} & ICCV'21 & OHG & 27.43 & 53.38 & 3.68 & 392M \\
RULSTM \cite{kini2023meccano} & CVIU'23 & OHG & 24.08 & 58.23 & 2.47 & 67M \\
VLMAH \cite{manousaki2023vlmah} & ICCV-W'23 & OHG & 28.90 & 58.13 & 0.72 & 45M \\
MMTF-RU \cite{mehta2025mmtfru} & T-ASE'25 & OHG & \textbf{29.75} & \textbf{64.46} & -- & 33.2M \\
AAG \cite{benavent2026aag} & WACV'26 & RGB+Depth & 27.21 & 51.15 & 0.71 & 24M \\
AAG+ \cite{benavent2026aagplus} & CoRR'26 & RGB+Depth+AH & 27.24 & 60.41 & -- & 34M \\
\midrule
\methodname & -- & Gaze(RGB + Hands) & 25.20 & 61.67 & \textbf{4.20} & 15M \\
\bottomrule
\end{tabular}%
}
\noindent\footnotesize
OHG: object, hand, and gaze features; AH: action history.
\end{table}

In terms of efficiency, our model uses 105M trainable parameters, which is more compact than ensemble-based approaches~\cite{kini2023meccano,bianchi2024egocentric} and single model approaches~\cite{shiota2024egocentric}, though larger than certain baselines~\cite{kini2023meccano, benavent2026aag}. Overall, these results show that graph-based representations can achieve competitive recognition performance with fewer parameters than several video-based approaches.

\begin{table}[t] 
\centering 
\scriptsize
\caption{Full graphs (FG) versus \methodname graphs efficiency statistics. EGTEA values denote averages over the three official splits.} 
\label{tab:graph_efficiency}
\setlength{\tabcolsep}{3pt} 
\begin{tabular}{llcccc} 
\toprule Dataset & Graph & Nodes $\downarrow$ & Edges $\downarrow$ & Avg.\ max distance $\downarrow$ & Global Efficiency $\uparrow$ \\ 
\midrule 
MECCANO & FG & 13.57 & 8.13 & 3.29 & 0.257 \\ & \methodname & \textbf{3.75} & \textbf{2.75} & \textbf{2.00} & \textbf{0.771} \\ \midrule EGTEA & FG & 4.51 & 3.41 & 2.03 & 0.703 \\ & \methodname & \textbf{3.77} & \textbf{2.77} & \textbf{2.00} & \textbf{0.769} \\ 
\bottomrule 
\end{tabular} 
\end{table}

\begin{figure}[h!]
    \centering
    \includegraphics[width=.9\linewidth]{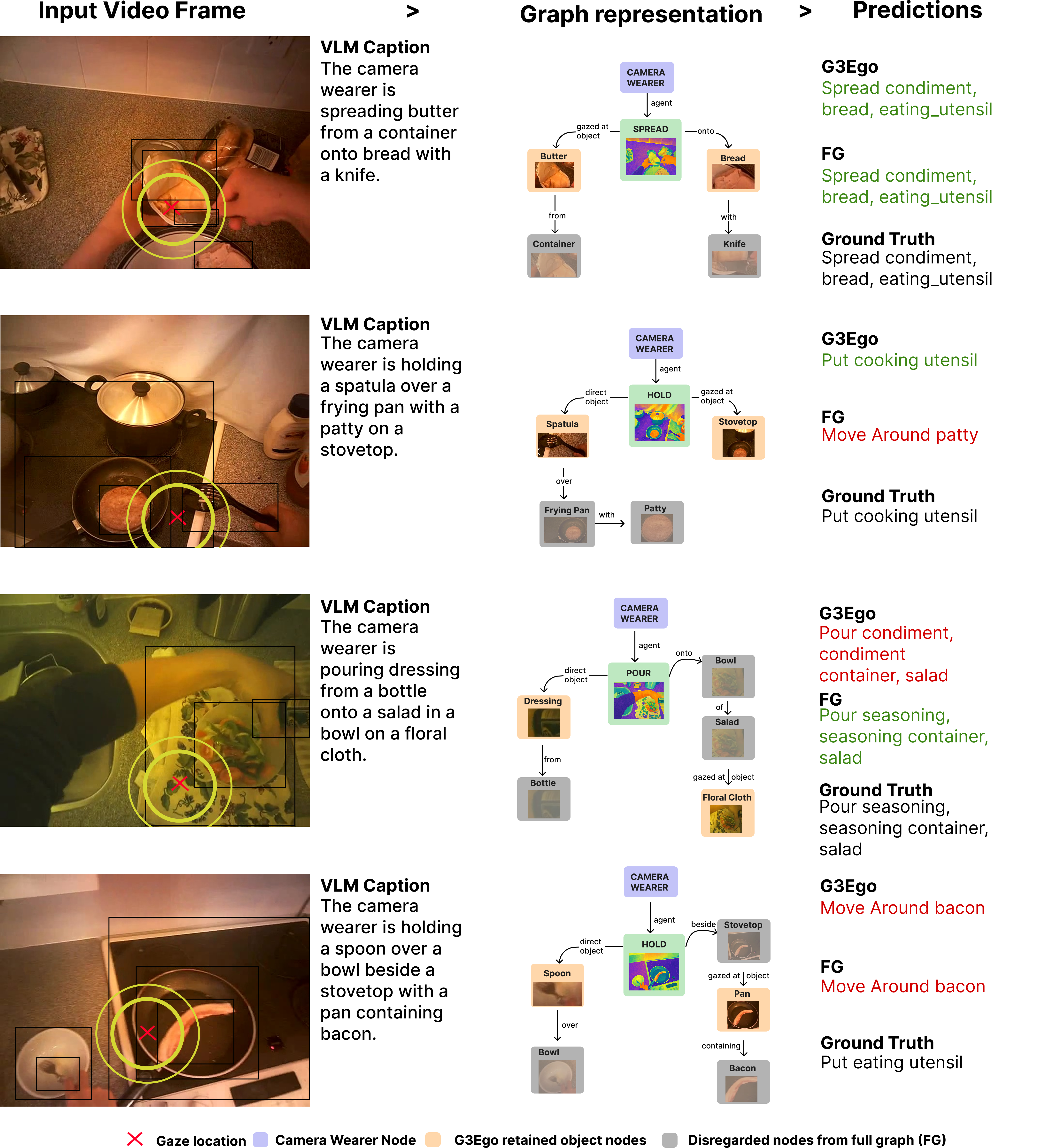}
    \caption{Qualitative examples of action classification predictions with the ground-truth labels. For each sample, we visualize the full action scene graph (FG) and the corresponding gaze-guided graph. Gaze-guided pruning preserves action-relevant entities and relationships while removing irrelevant context, yielding more discriminative graph representations and improved action recognition.}
    \label{fig:placeholder}
\end{figure}

\paragraph{\textbf{Action recognition on EGTEA Gaze+.}}
The results on EGTEA Gaze+ in Table \ref{tab:egtea_results} show that \methodname performs well beyond MECCANO. \methodname achieves the best Split 1 and Split 2 mean accuracies at 61.68 and 56.34, respectively. Although Split 3 remains lower at 51.46, the average mean accuracy across three splits is the highest for our model at 56.49. We also compare our method to works that report average Top-1 accuracy instead of mean accuracy across splits, and \methodname achieves 64.25, second only to GC-TSM \cite{hao2022group}. For a fair comparison, Table \ref{tab:egtea_results} includes only methods that do not rely on exocentric pretraining, while the complete comparison with all reported methods, including those that rely on exocentric pretraining, is provided in the supplementary material.
These results demonstrate that \methodname remains effective across different egocentric activity domains.

\paragraph{\textbf{Action anticipation performance on MECCANO.}}
We also evaluate the extension of our framework for next-activity prediction in MECCANO, reported in Table \ref{tab:meccano_reported_graph_results}. Our \methodname model achieves 25.20 Top-1, 61.67 Top-5, and 4.20 Macro-F1 at  $t_a=1$  before the next action. Although Top-1 accuracy is lower than the many baseline methods, it is important to note that the majority class baseline already achieves a 27.21 Top-1, which is comparable to many state-of-the-art methods in the table. This indicates that Top-1 is not a sufficiently informative metric for MECCANO anticipation, likely due to the long-tailed class distribution. In contrast, our method achieves the highest Macro-F1 among all reported approaches, suggesting a much stronger balance across classes and better performance on underrepresented actions.

\paragraph{\textbf{\methodname graph efficiency analysis.}}
Table~\ref{tab:graph_efficiency} shows how graph pruning affects graph efficiency on both MECCANO and EGTEA Gaze+. \methodname substantially reduces the size of the graph while producing shorter communication paths and higher global efficiency. In MECCANO, \methodname reduces the average number of nodes and edges by $72.4\%$ and $66.1\%$, respectively, while increasing the global efficiency from $0.257$ to $0.771$. A similar trend is observed on EGTEA Gaze+.

It is also notable that MECCANO contains substantially more nodes and edges than EGTEA Gaze+. This difference is caused by the longer captions generated by the Vision-Language Model, which tend to mention many objects present while the participant is assembling the toy motorbike. Consequently, graph pruning is particularly valuable for MECCANO, as it removes irrelevant objects and focuses the graph on the objects involved in the interaction.
\section{Conclusion}

By leveraging gaze as a structural pruning mechanism, \methodname produces a compact and interpretable interaction graph that models the camera wearer’s interactions with the surrounding scene. Experimental results on MECCANO and EGTEA Gaze+ demonstrate that \methodname achieves competitive performance across different egocentric benchmarks, while obtaining the strongest Macro-F1 results among compared methods on MECCANO and average mean accuracy across 3 splits on EGTEA Gaze+. These findings indicate that gaze-guided graph representations can effectively capture semantic action structure without requiring dense video pretraining.

Our analysis also highlights the importance of evaluating egocentric action understanding beyond accuracy-based metrics, particularly in long-tailed settings where majority classes can dominate performance. Future work will extend \methodname to model longer temporal dependencies, richer hand--object and human--object interactions, and more expressive graph representations that capture higher-order semantic relationships in egocentric videos.

\section{Acknowledgements}
E. Talavera Martinez was supported by the NWO Talent Programme – VENI (project Understanding Social Interactions in First-Person Videos with Multimodal Learning, file number 244507) which is financed by the Dutch Research Council (NWO).  Computational resources were provided by the University of Twente High-Performance Computing infrastructure.

\bibliographystyle{splncs04}
\bibliography{main}

\appendix

\clearpage
\begin{center}
    {\Large \textbf{\methodname: Gaze-Guided Graphs for Egocentric Action Understanding}}\\[0.5em]
    {\large Supplementary Material}
\end{center}

\FloatBarrier

\section{Model Card}
\label{app:model_card}

We use frozen Qwen3-VL-32B-Instruct~\cite{bai2025qwen3vl} for frame captioning at $448{\times}448$ resolution with 160 output tokens, spaCy~\cite{honnibal2020spacy} for parsing, DINOv3 ViT-L/16~\cite{dinov3_2023} for 1024-D global features, and GroundingDINO Swin-T~\cite{groundingdino2023} for 256-D object features and 20-D hand descriptors. Only the graph embedder and temporal aggregation module are trained. Grounding and training use A40 (40\,GB) or Quadro RTX 6000 (24\,GB) GPUs; VLM inference uses one RTX PRO 6000 Blackwell (98\,GB).

We distinguish between two computational stages. Frozen visual feature extraction and graph construction are performed offline and cached, whereas the trainable graph and temporal modules operate directly on the cached representations. We therefore report both the one-time
feature-construction cost and the downstream inference cost over cached graphs.

\begin{table}[t]
\centering
\scriptsize
\setlength{\tabcolsep}{2.2pt}
\renewcommand{\arraystretch}{0.95}

\caption{ Computational metrics of the frozen feature extraction components and the trainable temporal model. Latency is measured per frame for the frozen visual components and per graph sequence for the temporal model. Peak memory denotes measured peak allocated GPU memory.}
\label{tab:component_compute}

\resizebox{\columnwidth}{!}{%
\begin{tabular}{lcccc}
\toprule
\textbf{Component} & \textbf{Output} & \textbf{Params} & \textbf{GFLOPs} & \textbf{Memory} \\
\midrule
Qwen3-VL-32B & 160 tokens & 33.36B & 27,500.8 & 63.91\,GiB \\
DINOv3 ViT-L/16 & 1024-D & 303.13M & 121.76 & 1,176.52\,MiB \\
GroundingDINO & 256-D & 232.90M & 1,312.93 & 2,160.50\,MiB \\
TempAgg, pruned & 352-D & 105M & 0.435 & 421.84\,MiB \\
TempAgg, full & 352-D & 105M  & 0.47 & 422.44\,MiB \\
\bottomrule
\end{tabular}%
}
\end{table}

\section{Vision-Language Model Action and Activity Annotation Prompts}
\label{app:vlm_prompts}

\lstdefinestyle{prompt}{
  basicstyle=\ttfamily\scriptsize,
  breaklines=true,
  breakatwhitespace=true,
  columns=fullflexible,
  keepspaces=true,
  frame=single,
  framerule=0.3pt,
  rulecolor=\color{black!35},
  backgroundcolor=\color{black!2},
  xleftmargin=3pt,
  xrightmargin=3pt,
  aboveskip=4pt,
  belowskip=4pt,
  showstringspaces=false
}

\begin{lstlisting}[style=prompt]
You will analyze ONE egocentric image and output the camera wearer's action in the frame you are observing.

Output one sentence only, following this structure:

The camera wearer is <verb-ing> <object> <optional-context>.

Rules:
- Start with exactly: "The camera wearer is"
- Use a clear verb in the -ing form, such as opening, closing, holding, cutting, using, pushing, pulling, taking, placing, stirring, pouring, washing, peeling, spreading, mixing, serving, eating, drinking, operating, reaching, or looking.
- Identify the main object being handled or attended to.
- Focus on the camera wearer's hands and the objects with which they interact.
- Add short scene context only when it helps identify the action or relevant objects.
- Include object attributes only when needed to distinguish the object.
- Associate every attribute with a specific object.
- Mention other people when they are involved in the action.
- If no clear hand--object interaction is visible, describe the most likely action supported by the scene.
- Be specific about visible objects, but do not explain the reasoning.

Examples:
The camera wearer is opening a cabinet above a counter.
The camera wearer is cutting a vegetable on a cutting board.
The camera wearer is pouring liquid into a cup beside a sink.
The camera wearer is holding a bowl over a counter.
The camera wearer is reaching for a bottle.
The camera wearer is playing table tennis with another person.

Return only the sentence. Do not explain the reasoning.
\end{lstlisting}

We use Qwen3-VL-32B-Instruct~\cite{bai2025qwen3vl} in mixed precision without quantization. Input frames are resized to $448 \times 448$ pixels and encoded using the Hugging Face chat template. We use greedy decoding with \texttt{max\_new\_tokens=160}, and the default \texttt{num\_beams=1}. Temperature and top-$p$ sampling are not applied because sampling is disabled.

Each frame is annotated independently. The prompt constrains the output to a single sentence and emphasizes the camera wearer’s hands and manipulated objects as the primary evidence for the action.

\section{Additional Component Ablations}
\label{app:additional_comp_ablation}

We investigate whether the semantic information produced by the VLM alone is sufficient for action recognition, without the structured graph representation proposed in \methodname. To this end, we use Qwen3-VL-32B-Instruct~\cite{bai2025qwen3vl} to generate frame-level scene descriptions, which in our full pipeline are subsequently parsed, visually grounded, and converted into action scene graphs. For this ablation, however, we discard the graph construction stage and instead encode the generated captions using CLIP~\cite{radford2021learning}. An LSTM is then trained over the resulting sequence of text embeddings for action recognition. As shown in Table~\ref{tab:additional_ablation}, this caption-only variant performs below the majority-class baseline on MECCANO~\cite{kini2023meccano}. These results indicate that caption embeddings alone fail to capture the temporal dynamics and fine-grained interaction cues required for reliable action recognition, highlighting that the performance gains of \methodname stem from its structured graph representation rather than from the VLM-generated captions themselves.

We also evaluate whether the improvements from gaze-guided pruning are simply a consequence of reducing graph complexity and retaining fewer objects. We therefore introduce a random-pruning baseline that retains two randomly selected object nodes per frame, matching the object-node budget of \methodname. As shown in Table~\ref{tab:additional_ablation}, random pruning performs worse than both the full graph and gaze-guided pruning. This result indicates that the benefit of \methodname is not explained by sparsification alone, but by the use of gaze to preserve action-relevant objects.

Finally, we examine whether gaze is beneficial when used as an explicit auxiliary input in addition to its role in graph pruning. To this end, we augment \methodname with frame-level gaze coordinates as an auxiliary modality. As reported in Table~\ref{tab:additional_ablation}, adding gaze coordinates as an auxiliary feature reduces performance. One possible explanation is that frame-level gaze coordinates are noisy and vary rapidly as the camera wearer shifts attention across consecutive frames. In this setting, due to noise directly providing gaze as an additional modality may distract the temporal model, whereas using gaze for graph pruning yields a more stable representation.

\begin{table}[!t]
\centering
\scriptsize
\setlength{\tabcolsep}{1.7pt}
\renewcommand{\arraystretch}{1.08}

\caption{
Additional ablations on MECCANO action recognition. All variants use an LSTM over the same 10 uniformly sampled  frames and exclude hand features. VLM caption embeddings are obtained using the CLIP~\cite{radford2021learning} text encoder. Random-2 retains two randomly selected object nodes per frame. Gaze Aux.\ indicates that gaze
coordinates are explicitly provided as an additional  recognition feature.
}
\label{tab:additional_ablation}

\begin{tabular*}{\columnwidth}{
    @{\extracolsep{\fill}}
    >{\raggedright\arraybackslash}p{0.20\columnwidth}
    c c c c c c c
    @{}
}
\toprule
\makecell[c]{\textbf{Input}\\\textbf{Representation}}
& \makecell{\textbf{Object}\\\textbf{Selection}}
& \makecell{\textbf{Gaze}\\\textbf{Aux.}}
& \makecell{\textbf{Temporal}\\\textbf{Model}}
& \makecell{\textbf{N}\\\textbf{Frames}}
& \textbf{Top-1}
& \textbf{Top-5}
& \textbf{F1} \\
\midrule

\multicolumn{8}{c}{\itshape
Effect of VLM Semantic Prior} \\[1pt]
Majority class baseline
& -- & \xmark & -- & --
& 27.31 & 51.22 & 0.72 \\

\makecell[l]{VLM caption\\CLIP~\cite{radford2021learning} embeddings}
& -- & \xmark & LSTM & 10
& 3.86 & 24.76 & 0.46 \\

\multicolumn{8}{c}{\itshape
Effect of Pruning Strategy} \\[1pt]

FG
& -- & \xmark & LSTM & 10
& 37.34  & 70.39 & 8.61 \\

\methodname
& Random-2 & \xmark & LSTM & 10
& 21.22 & 57.07 & 7.95 \\

\methodname
& Gaze pruning & \xmark & LSTM & 10
& \textbf{37.90} & \textbf{70.88} & \textbf{10.63} \\

\midrule
\multicolumn{8}{c}{\itshape Effect of Explicit Gaze Input} \\[1pt]

\methodname
& Gaze pruning & \cmark & LSTM & 10
& 23.45 & 57.81 & 8.62 \\

\methodname
& Gaze pruning & \xmark & LSTM & 10
& \textbf{37.90} & \textbf{70.88} & \textbf{10.63} \\

\bottomrule
\end{tabular*}
\end{table}

\section{Additional results on EGTEA Gaze+}
\label{app:egtea_add_results}

In Table~\ref{tab:egtea_full_results}, we provide a broader comparison with methods using different pretraining regimes. The upper part of the table reports approaches directly comparable to \methodname, as they do not rely on exocentric video pretraining, while the lower part includes methods initialized from large-scale video datasets, such as Kinetics~\cite{carreira2017quo} or EPIC-KITCHENS-100~\cite{damen2022epickitchens100}, before being fine-tuned on EGTEA Gaze+. These settings are not directly comparable because the pretrained video models benefit from task-relevant spatio-temporal representations and substantially greater pretraining supervision.

In contrast, \methodname does not use a video-pretrained backbone or egocentric-domain pretraining. Its global frame representation is extracted using the general-purpose vision encoder DINOv3~\cite{dinov3_2023}, and temporal information is modeled only after the frame-level graph representations have been constructed. Despite the absence of video-specific pretraining, \methodname remains competitive with several pretrained approaches, although a gap remains to the strongest methods that use Kinetics or EPIC-KITCHENS-100 pretraining and additional annotated hand--object or object-state supervision. The results suggest that \methodname recovers part of the benefits provided by video pretraining while operating on sparsely sampled frame-level representations and without requiring the additional supervision used by those models, while remaining the strongest method among approaches that do not rely on any dense video pretraining.

\begin{table*}[!t]
\centering
\scriptsize
\setlength{\tabcolsep}{2.2pt}
\renewcommand{\arraystretch}{0.92}
\caption{
EGTEA Gaze+ action-recognition results on the official three splits. S1--S3 report mAcc when available; otherwise, they report per-split Top-1 or overall accuracy. The upper block contains methods without
exocentric video pretraining, while the lower block contains methods using Kinetics~\cite{carreira2017quo} or EPIC-KITCHENS-100~\cite{damen2022epickitchens100} pretraining. Results across these blocks are provided for context and are not directly comparable because of differences in pretraining data, backbone design, and supervision.
}
\label{tab:egtea_full_results}
\resizebox{\textwidth}{!}{%
\begin{tabular}{@{}llp{1.8cm}p{3.0cm}ccccc@{}}
\toprule
\textbf{Method} & \textbf{Venue} & \textbf{Pretraining} &
\textbf{Modality / Supervision} &
\textbf{S1} & \textbf{S2} & \textbf{S3} &
\textbf{Avg mAcc} & \textbf{Avg Top-1} \\
\midrule

Two Stream~\cite{simonyan2014twostream}
& NeurIPS'14 & -- & RGB + Optical Flow
& 43.78 & 41.47 & 40.28 & 41.84 & -- \\

EgoIDT+Gaze~\cite{EgoIDT2015CVPR}
& CVPR'15 & -- & RGB + IDT + Gaze
& 42.55 & 37.30 & 37.60 & 39.13 & -- \\

I3D+EgoConv~\cite{DeepDescriptors2016CVPR}
& CVPR'16 & -- & RGB + Hands + Head Motion + Gaze
& 54.19 & 51.45 & 49.41 & 51.68 & -- \\

TSN~\cite{wang2016temporal}
& ECCV'16 & -- & RGB + Optical Flow
& 58.01 & 55.01 & 54.78 & 55.93 & -- \\

Sudhakaran et al.~\cite{sudhakaran2018attention}
& ECCVW'18 & -- & RGB
& 52.40 & 50.09 & 49.11 & 50.53 & -- \\

LSTA~\cite{sudhakaran2019lsta}
& CVPR'19 & -- & RGB + Optical Flow
& 53.00 & -- & -- & -- & -- \\

Mutual Context~\cite{mutual_context2020huang}
& TIP'20 & -- & RGB + Multi-task Supervision
& 55.70 & -- & -- & -- & 62.60$^{\ddagger}$ \\

I3D~\cite{wang2021ipl}
& ICCV'21 & -- & RGB + Separate Classifiers
& 56.78 & 54.92 & 53.94 & 55.21 & -- \\

GC-TSM~\cite{hao2022group}
& CVPR'22 & -- & RGB
& -- & -- & -- & -- & 65.10 \\

I3D+Gaze~\cite{Li2023Eye}
& TPAMI'23 & -- & RGB + Gaze Pooling
& 53.74 & 50.30 & 49.63 & 51.22 & -- \\

Prob-ATT w/ Gaze~\cite{Li2023Eye}
& TPAMI'23 & -- & RGB + Gaze Attention
& 57.20 & 53.75 & 54.13 & 55.03 & -- \\

\midrule

I3D Joint~\cite{carreira2017quo}
& CVPR'17 & Kinetics~\cite{carreira2017quo} & RGB
& 55.76 & 53.14 & 53.55 & 54.15 & -- \\

STAM 2-Stream~\cite{lu2019stam}
& ICCVW'19 & Kinetics~\cite{carreira2017quo} & RGB + Optical Flow
& 60.54 & 55.21 & 55.32 & 57.02 & 65.97 \\

Multitask~\cite{kapidis2019multitask}
& ICCVW'19 & Kinetics~\cite{carreira2017quo} & RGB + Sub-task Supervision
& 61.40 & -- & -- & 57.60 & 65.70 \\

IPL I3D~\cite{wang2021ipl}
& ICCV'21 & Kinetics~\cite{carreira2017quo} & RGB
& 60.15 & \underline{59.03} & 57.98 & 59.05 & -- \\

SlowFast+HOCL+OSL~\cite{shiota2024egocentric}
& WACV'24 & Kinetics~\cite{carreira2017quo} & RGB + HOCL + OSL
& 59.36 & 57.39 & 57.66 & 58.14 & 66.86 \\

Swin-B+OSL~\cite{shiota2024egocentric}
& WACV'24 & EPIC-100~\cite{damen2022epickitchens100} & RGB + OSL
& 65.89 & \textbf{63.96} & \underline{59.69} &
\textbf{63.18} & \textbf{69.65} \\

MACS-ViT (L)~\cite{lu2025mixed}
& TOMM'25 & Kinetics~\cite{carreira2017quo} & RGB
& -- & -- & -- & -- & 67.30$^{\ddagger}$ \\

\midrule
\methodname
& -- & -- & Gaze(RGB + Hands) & 61.68 & 56.34 
& 51.46 & 56.49 & 64.25 \\
\bottomrule
\end{tabular}%
}

\vspace{0.4mm}
\raggedright
\scriptsize
$^{\ddagger}$Overall Top-1 accuracy reported on a single split or without split-wise mAcc; therefore, it is not directly comparable to the average split-wise mAcc values.\par
\vspace{0.2em}
\noindent
\methodname: gaze-guided graphs for egocentric action understanding; IDT: improved dense trajectories;
HOCL: hand--object contact learning; OSL: object-state learning.
\end{table*}

\end{document}